\documentclass{article} 
\pdfoutput=1
\usepackage{nips13submit_e,times}
\usepackage{hyperref}
\usepackage{url}

\usepackage{amsmath}
\usepackage{amssymb}
\usepackage{amsfonts}
\usepackage{stmaryrd}

\usepackage{graphicx}
\usepackage{wrapfig}

\let\uaccent\u

\newcommand{\T}{\mathrm{T}}

\newcommand{\V}{\mathbf{V}}
\renewcommand{\v}{\mathbf{v}}
\newcommand{\U}{\mathbf{U}}
\renewcommand{\u}{\mathbf{u}}
\newcommand{\h}{\mathbf{h}}
\newcommand{\I}{\mathbf{I}}
\renewcommand{\k}{\mathbf{k}}
\newcommand{\K}{\mathbf{K}}
\newcommand{\w}{\mathbf{w}}
\newcommand{\W}{\mathbf{W}}

\title{Predicting Parameters in Deep Learning}

\author{
Misha Denil$^1$ \quad
Babak Shakibi$^2$ \quad
Laurent Dinh$^3$ \\
\textbf{Marc'Aurelio Ranzato}$^4$ \quad
\textbf{Nando de Freitas}$^{1,2}$ \\
$^1$University of Oxford, United Kingdom \\
$^2$University of British Columbia, Canada \\
$^3$Universit\'e de Montr\'eal, Canada \\
$^4$Facebook Inc., USA \\
\texttt{\{misha.denil,nando.de.freitas\}@cs.ox.ac.uk} \\
\texttt{laurent.dinh@umontreal.ca} \\
\texttt{ranzato@fb.com}
}

\usepackage{todonotes}

\nipsfinalcopy 

\begin{document}

\maketitle

\begin{abstract}
  We demonstrate that there is significant redundancy in the parameterization of
  several deep learning models.  Given only a few weight values for each feature
  it is possible to accurately predict the remaining values.  Moreover, we show
  that not only can the parameter values be predicted, but many of them need not
  be learned at all.  We train several different architectures by learning only
  a small number of weights and predicting the rest.  In the best case we are
  able to predict more than 95\% of the weights of a network without any drop in
  accuracy.
\end{abstract}

\section{Introduction}

Recent work on scaling deep networks has led to the construction of the largest
artificial neural networks to date.  It is now possible to train networks with
tens of millions~\cite{alex_imagenet} or even over a billion
parameters~\cite{dean2012, quoc2012}.

The largest networks (i.e.\ those of Dean \emph{et al.}~\cite{dean2012}) are
trained using asynchronous SGD.  In this framework many copies of the model
parameters are distributed over many machines and updated independently.  An
additional synchronization mechanism coordinates between the machines to ensure
that different copies of the same set of parameters do not drift far from each
other.

A major drawback of this technique is that training is very inefficient in how
it makes use of parallel resources~\cite{bengio2013}.  In the largest networks
of Dean \emph{et al.}~\cite{dean2012}, where the gains from distribution are
largest, distributing the model over 81 machines reduces the training time per
mini-batch by a factor of 12, and increasing to 128 machines achieves a speedup
factor of roughly 14.  While these speedups are very significant, there is a
clear trend of diminishing returns as the overhead of coordinating between the
machines grows.  Other approaches to distributed learning of neural networks
involve training in batch mode~\cite{deng2012scalable}, but these methods have
not been scaled nearly as far as their online counterparts.

It seems clear that distributed architectures will always be required for
extremely large networks; however, as efficiency decreases with greater
distribution, it also makes sense to study techniques for learning larger
networks on a single machine.  If we can reduce the number of parameters which
must be learned and communicated over the network of fixed size, then we can
reduce the number of machines required to train it, and hence also reduce the
overhead of coordination in a distributed framework.

In this work we study techniques for reducing the number of free parameters in
neural networks by exploiting the fact that the weights in learned networks tend
to be structured.  The technique we present is extremely general, and can be
applied to a broad range of models.  Our technique is also completely orthogonal
to the choice of activation function as well as other learning optimizations; it
can work alongside other recent advances in neural network training such as
dropout~\cite{hinton_dropout}, rectified units~\cite{nair2010rectified} and
maxout~\cite{goodfellow_maxout} without modification.

\begin{figure}[t]
  \centering
  \includegraphics[width=0.15\linewidth]{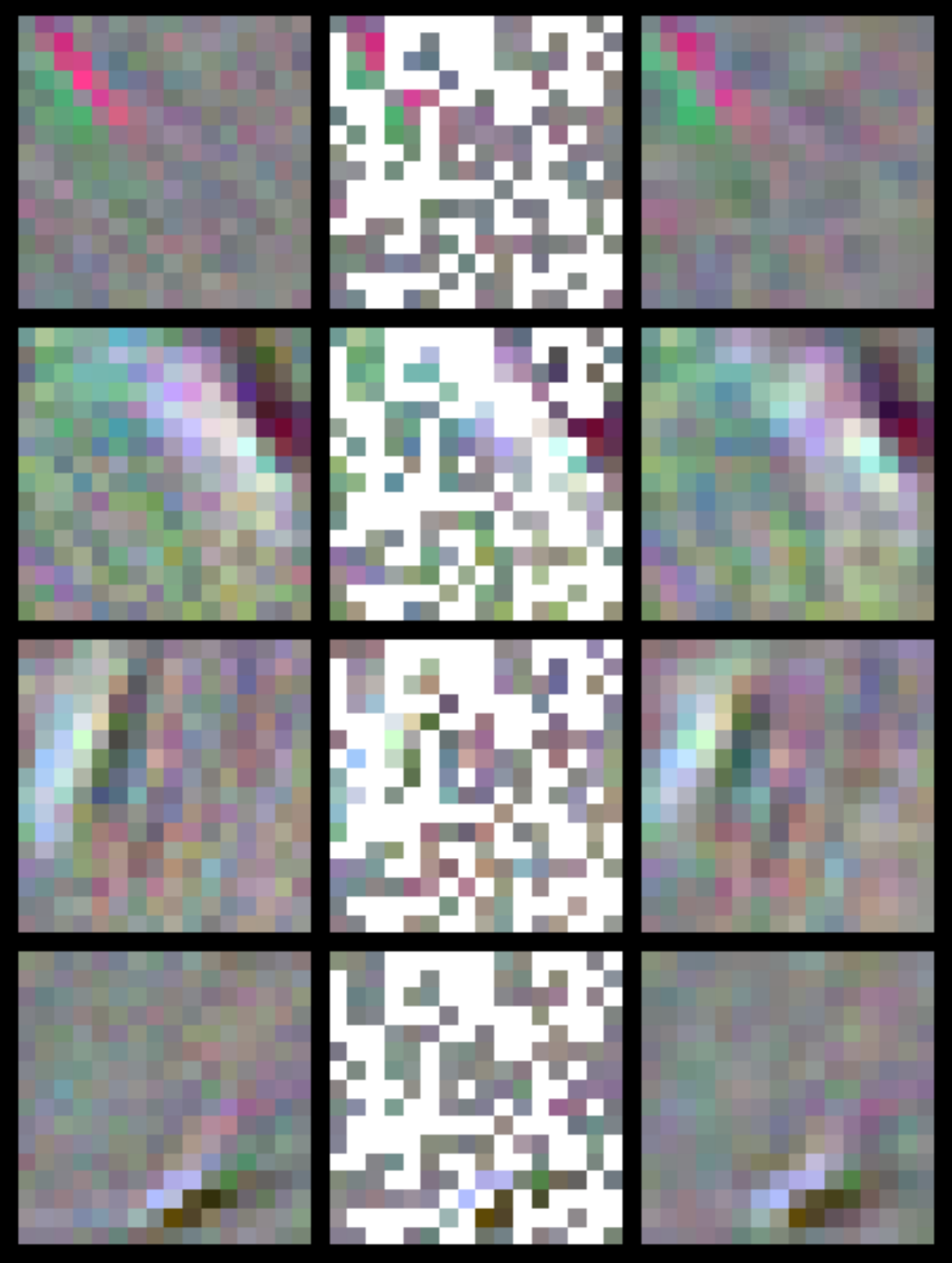}
  \qquad
  \includegraphics[width=0.15\linewidth]{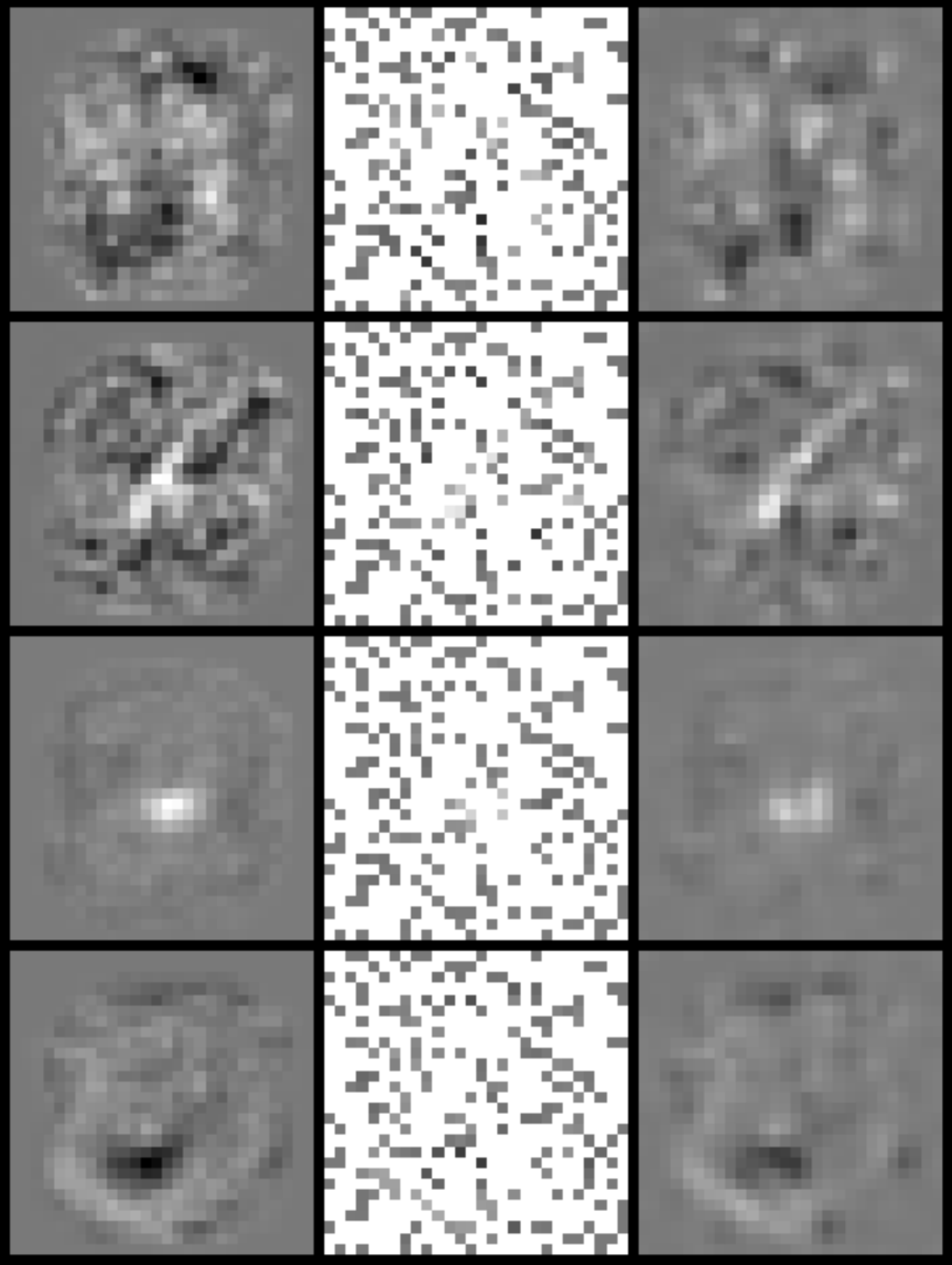}
  \qquad
  \includegraphics[width=0.15\linewidth]{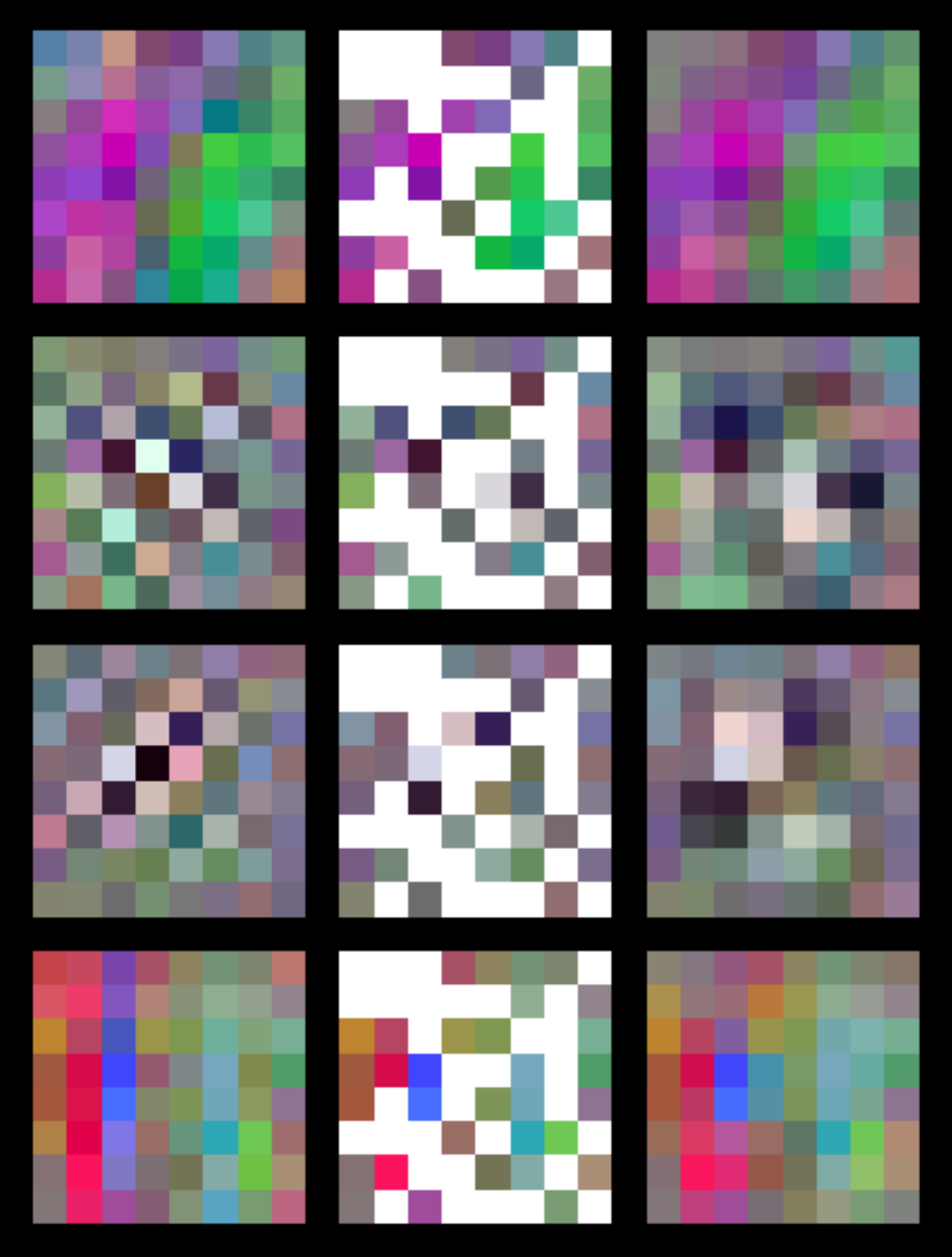}
  \qquad
  \includegraphics[width=0.15\linewidth]{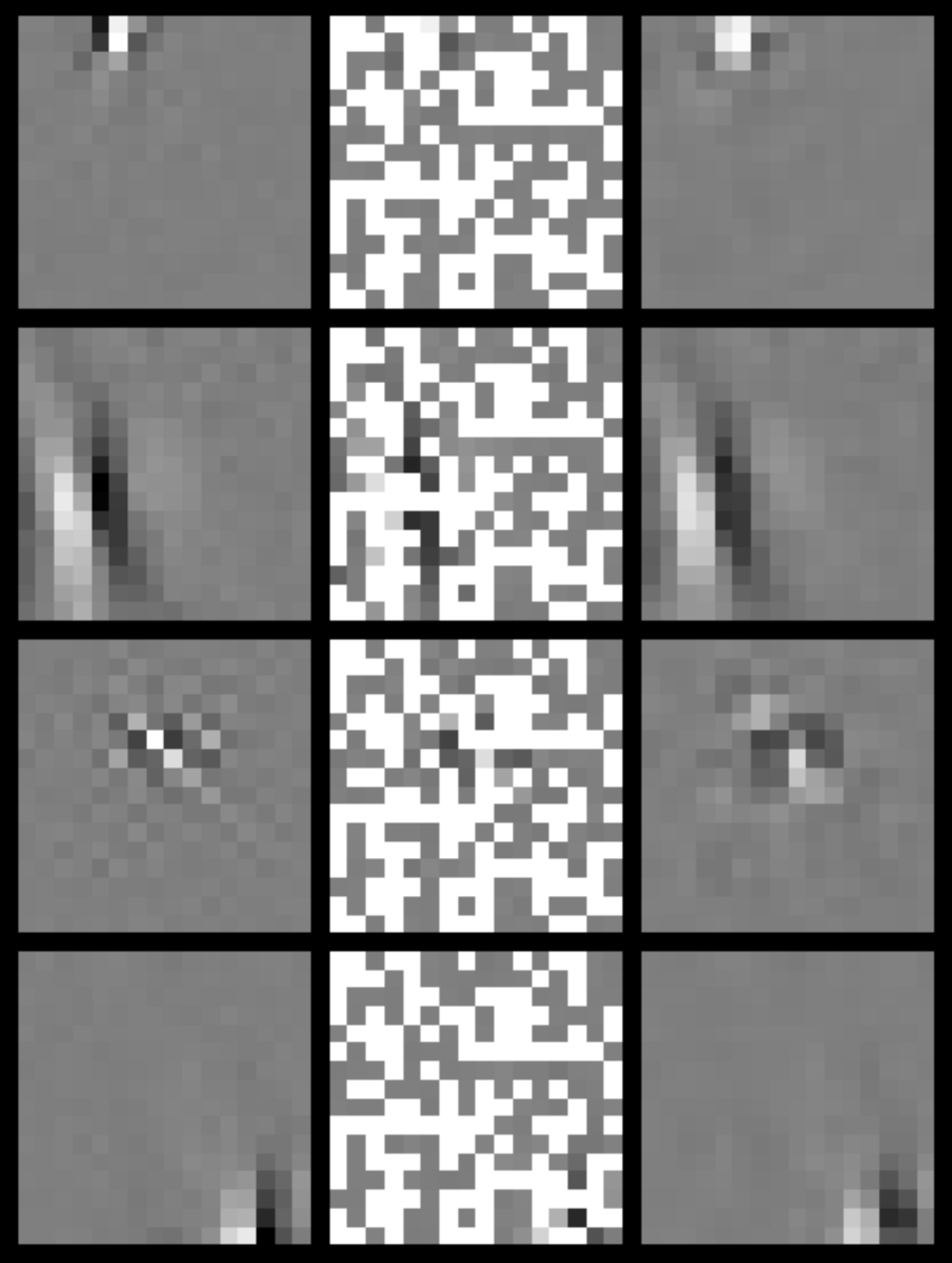}
  \qquad
  \includegraphics[width=0.15\linewidth]{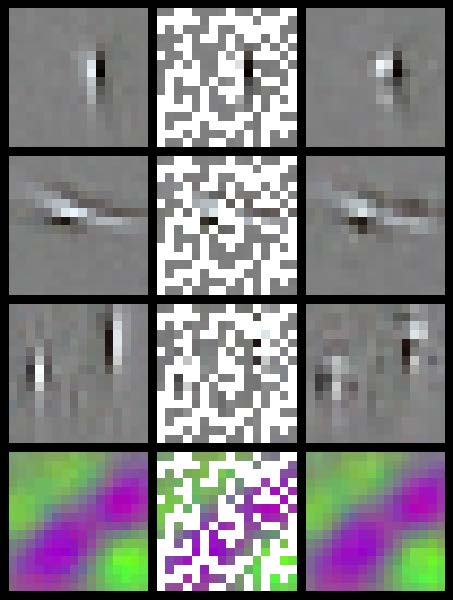}
  \caption{The first column in each block shows four learned features
    (parameters of a deep model). The second column shows a few parameters
    chosen at random from the original set in the first column. The third column
    shows that this random set can be used to predict the remaining parameters.
    From left to right the blocks are: (1) a convnet trained on STL-10 (2) an
    MLP trained on MNIST, (3) a convnet trained on CIFAR-10, (4) Reconstruction
    ICA trained on Hyv\"arinen's natural image dataset (5) Reconstruction ICA
    trained on STL-10.}
  \label{fig:feature-reconstruction}
\end{figure}

\begin{wrapfigure}{r}{.4\textwidth}
  \centering
  \vspace{-8pt}
  \includegraphics[width=0.4\textwidth]{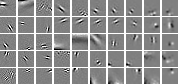}
  \caption{RICA with different amounts of parameter prediction.  In the leftmost
    column 100\% of the parameters are learned with L-BFGS. In the rightmost
    column, only 10\% of the parameters learned, while the remaining values are
    predicted at each iteration.  The intermediate columns interpolate between
    these extremes in increments of 10\%.}
   \label{fig:feature-reconstruction2}   
\end{wrapfigure}

The intuition motivating the techniques in this paper is the well known
observation that the first layer features of a neural network trained on natural
image patches tend to be globally smooth with local edge features, similar to
local Gabor features~\cite{coates_analysis, alex_imagenet}.  Given this
structure, representing the value of each pixel in the feature separately is
redundant, since it is highly likely that the value of a pixel will be equal to
a weighted average of its neighbours.  Taking advantage of this type of
structure means we do not need to store weights for every input in each
feature. This intuition is illustrated in
Figures~\ref{fig:feature-reconstruction} and~\ref{fig:feature-reconstruction2}.

The remainder of this paper is dedicated to elaborating on this observation.  We
describe a general purpose technique for reducing the number of free parameters
in neural networks.  The core of the technique is based on representing the
weight matrix as a low rank product of two smaller matrices.  By factoring the
weight matrix we are able to directly control the size of the parameterization
by controlling the rank of the weight matrix.

Na\"ive application of this technique is straightforward but tends to reduce
performance of the networks.  We show that by carefully constructing one of the
factors, while learning only the other factor, we can train networks with vastly
fewer parameters which achieve the same performance as full networks with the
same structure.

The key to constructing a good first factor is exploiting smoothness in the
structure of the inputs.  When we have prior knowledge of the smoothness
structure we expect to see (e.g.\ in natural images), we can impose this
structure directly through the choice of factor.  When no such prior knowledge
is available we show that it is still possible to make a good data driven
choice.

We demonstrate experimentally that our parameter prediction technique is
extremely effective.  In the best cases we are able to predict more than 95\% of
the parameters of a network without any drop in predictive accuracy.


Throughout this paper we make a distinction between dynamic and static
parameters.  Dynamic parameters are updated frequently during learning,
potentially after each observation or mini-batch.  This is in contrast to static
parameters, whose values are computed once and not altered.  Although the values
of these parameters may depend on the data and may be expensive to compute, the
computation need only be done once during the entire learning process.

The reason for this distinction is that static parameters are much easier to
handle in a distributed system, even if their values must be shared between
machines.  Since the values of static parameters do not change, access to them
does not need to be synchronized.  Copies of these parameters can be safely
distributed across machines without any of the synchronization overhead incurred
by distributing dynamic parameters.


\section{Low rank weight matrices}

Deep networks are composed of several layers of transformations of the form $\h
= g(\v\W)$, where $\v$ is an $n_v$-dimensional input, $\h$ is an
$n_h$-dimensional output, and $\W$ is an $n_v \times n_h$ matrix of parameters.
A column of $\W$ contains the weights connecting each unit in the visible layer
to a single unit in the hidden layer.  We can to reduce the number of free
parameters by representing $\W$ as the product of two matrices $\W = \U\V$,
where $\U$ has size $n_v\times n_\alpha$ and $\V$ has size $n_\alpha\times n_h$.
By making $n_\alpha$ much smaller than $n_v$ and $n_h$ we achieve a substantial
reduction in the number of parameters.

In principle, learning the factored weight matrices is straightforward.  We
simply replace $\W$ with $\U\V$ in the objective function and compute
derivatives with respect to $\U$ and $\V$ instead of $\W$.  In practice this
na\"ive approach does not preform as well as learning a full rank weight matrix
directly.

Moreover, the factored representation has redundancy.  If $\mathbf{Q}$ is any
invertible matrix of size $n_\alpha \times n_\alpha$ we have $\W = \U\V =
(\U\mathbf{Q})(\mathbf{Q}^{-1}\V) = \tilde{\U}\tilde{\V}$.  One way to remove
this redundancy is to fix the value of $\U$ and learn only $\V$.  The question
remains what is a reasonable choice for $\U$?  The following section provides
an answer to this question.

\section{Feature prediction}
\label{sec:feature-prediction}

We can exploit the structure in the features of a deep network to represent the
features in a much lower dimensional space.  To do this we consider the weights
connected to a single hidden unit as a function $\w : \mathcal{W} \to
\mathbb{R}$ mapping weight space to real numbers estimate values of this
function using regression.  In the case of $p\times p$ image patches,
$\mathcal{W}$ is the coordinates of each pixel, but other structures for
$\mathcal{W}$ are possible.

A simple regression model which is appropriate here is a linear combination of
basis functions.  In this view the columns of $\U$ form a dictionary of basis
functions, and the features of the network are linear combinations of these
features parameterized by $\V$.  The problem thus becomes one of choosing a good
base dictionary for representing network features.




\subsection{Choice of dictionary}

The base dictionary for
feature prediction can be constructed in several ways.  An obvious choice is to train a single layer unsupervised
model and use the features from that model as a dictionary.  This approach has
the advantage of being extremely flexible---no assumptions about the structure
of feature space are required---but has the drawback of requiring an additional
training phase.

When we have prior knowledge about the structure of feature space we
can exploit it to construct an appropriate dictionary.  For example when
learning features for images we could choose $\U$ to be a selection of
Fourier or wavelet bases to encode our expectation of smoothness.

We can also build $\U$ using kernels that encode prior knowledge.  One way to achieve this is via kernel ridge regression \cite{Shawe-Taylor:2004}.  Let
$\w_\alpha$ denote the observed values of the weight vector $\w$ on a restricted
subset of its domain $\alpha \subset \mathcal{W}$.  We introduce a kernel matrix
$\K_\alpha$, with entries $(\K_\alpha)_{ij} = k(i,j)$, to model the covariance
between locations $i,j \in \alpha$.  The parameters at these locations are
$(\w_\alpha)_i$ and $(\w_\alpha)_j$.  The kernel enables us to make smooth
predictions of the parameter vector over the entire domain $\mathcal{W}$ using
the standard kernel ridge predictor:
\begin{align*}
  \w = \k_\alpha^\T(\K_\alpha + \lambda\I)^{-1}\w_\alpha \enspace,
\end{align*}
where $\k_\alpha$ is a matrix whose elements are given by $(\k_\alpha)_{ij} =
k(i,j)$ for $i\in\alpha$ and $j\in\mathcal{W}$, and $\lambda$ is a ridge
regularization coefficient.  In this case we have $\U
= \k_\alpha^\T(\K_\alpha + \lambda\I)^{-1}$ and $\V = \w_\alpha$.


\subsection{A concrete example}

In this section we describe the feature prediction process as it applies to
features derived from image patches using kernel ridge regression, since the
intuition is strongest in this case.  We defer a discussion of how to select a
kernel for deep layers as well as for non-image data in the visible layer to a
later section.  In those settings the prediction process is formally identical,
but the intuition is less clear.

If $\v$ is a vectorized image patch corresponding to the visible layer of a
standard neural network then the hidden activity induced by this patch is given
by $\h = g(\v\W)$, where $g$ is the network nonlinearity and $\W
= \begin{bmatrix} \w_1, \ldots, \w_{n_h} \end{bmatrix}$ is a weight matrix whose
columns each correspond to features which are to be matched to the visible
layer.

We consider a single column of the weight matrix, $\w$, whose elements are
indexed by $i \in \mathcal{W}$.  In the case of an image patch these indices are
multidimensional $i = (i_x, i_y, i_c)$, indicating the spatial location and
colour channel of the index $i$.  We select locations $\alpha\subset\mathcal{W}$
at which to represent the filter explicitly and use $\w_\alpha$ to denote the
vector of weights at these locations.

There are a wide variety of options for how $\alpha$ can be selected.  We have
found that choosing $\alpha$ uniformly at random from $\mathcal{W}$ (but tied
across channels) works well; however, it is possible that performance could be
improved by carefully designing a process for selecting $\alpha$.

We can use values for $\w_\alpha$ to predict the full feature as $\w =
\k_\alpha^\T(\K_\alpha+\lambda\I)^{-1}\w_\alpha$.  Notice that we can predict
the entire feature matrix in parallel using $\W =
\k_\alpha^\T(\K_\alpha+\lambda\I)^{-1}\W_\alpha$ where $\W_\alpha
= \begin{bmatrix} (\w_1)_\alpha, \ldots, (\w_{n_h})_\alpha \end{bmatrix}$.

For image patches, where we expect smoothness in pixel space, an appropriate
kernel is the squared exponential kernel
\begin{align*}
  k(i,j) = \exp\left(-\frac{(i_x - j_x)^2 + (i_y - j_y)^2}{2\sigma^2}\right)
\end{align*}
where $\sigma$ is a length scale parameter which controls the degree of
smoothness.

Here $\alpha$ has a convenient interpretation as the set of pixel locations in
the image, each corresponding to a basis function in the dictionary defined by
the kernel.  More generically we will use $\alpha$ to index a collection of
dictionary elements in the remainder of the paper, even when a dictionary
element may not correspond directly to a pixel location as in this example.

\subsection{Interpretation as pooling}

So far we have motivated our technique as a method for predicting features in a
neural network; however, the same approach can also be interpreted as a linear
pooling process.

Recall that the hidden activations in a standard neural network before applying
the nonlinearity are given by $g^{-1}(\h) = \v\W$.  Our motivation has proceeded
along the lines of replacing $\W$ with $\U_\alpha\W_\alpha$ and discussing the
relationship between $\W$ and its predicted counterpart.

Alternatively we can write $g^{-1}(\h) = \v_\alpha\W_\alpha$ where $\v_\alpha =
\v\U_\alpha$ is a linear transformation of the data.  Under this interpretation
we can think of a predicted layer as being composed to two layers internally.
The first is a linear layer which applies a fixed pooling operator given by
$\U_\alpha$, and the second is an ordinary fully connected layer with $|\alpha|$
visible units.

\subsection{Columnar architecture}
\label{ssec:columns}

The prediction process we have described so far assumes that $\U_\alpha$ is the
same for all features; however, this can be too restrictive.  Continuing with
the intuition that filters should be smooth local edge detectors we might want
to choose $\alpha$ to give high resolution in a local area of pixel space while
using a sparser representation in the remainder of the space.  Naturally, in
this case we would want to choose several different $\alpha$'s, each of which
concentrates high resolution information in different regions.

It is straightforward to extend feature prediction to this setting.  Suppose we
have several different index sets $\alpha_1, \ldots, \alpha_J$ corresponding to
elements from a dictionary $\U$.  For each $\alpha_j$ we can form the
sub-dictionary $\U_{\alpha_j}$ and predicted the feature matrix $\W_{j} =
\U_{\alpha_j}\W_{\alpha_j}$.  The full predicted feature matrix is formed by
concatenating each of these matrices blockwise $\W = \begin{bmatrix} \W_1,
  \ldots, \W_J \end{bmatrix}$.  Each block of the full predicted feature matrix
can be treated completely independently.  Blocks $\W_i$ and $\W_j$ share no
parameters---even their corresponding dictionaries are different.

\begin{figure}[tb]
  \centering
  \includegraphics[width=0.34\linewidth]{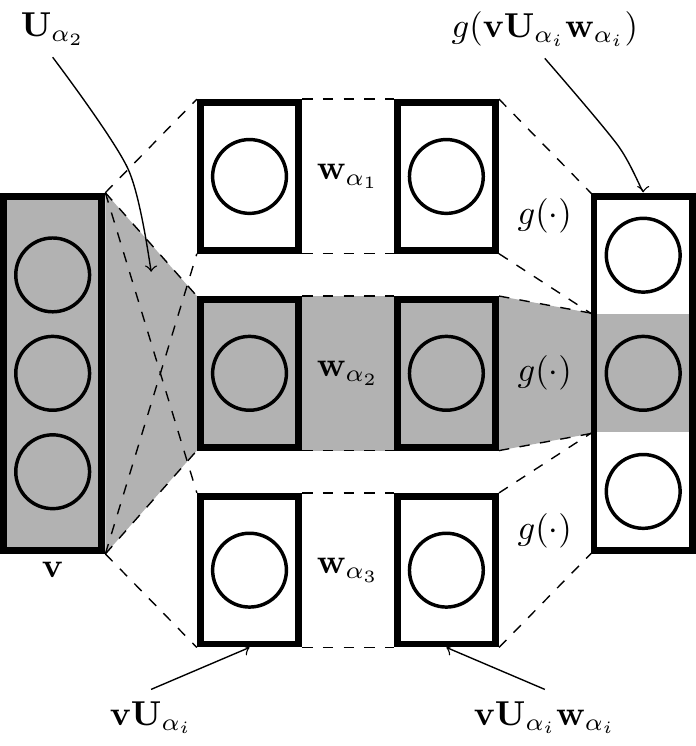}
  \hspace{2cm}
  \includegraphics[width=0.34\linewidth]{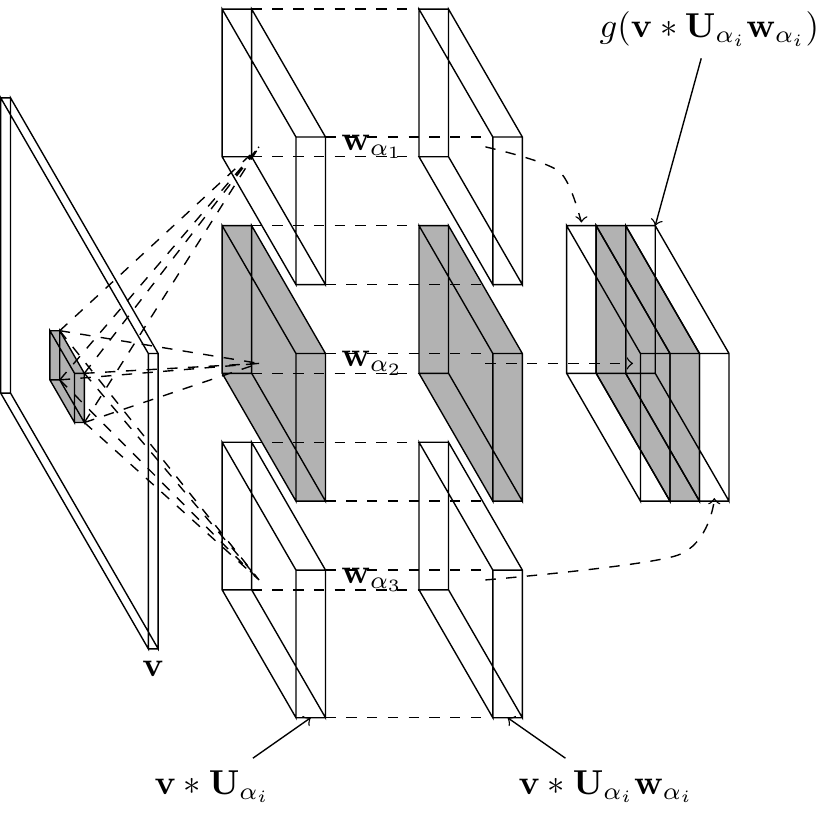}
  \caption{\textbf{Left:} Columnar architecture in a fully connected network,
    with the path through one column highlighted.  Each column corresponds to a
    different $\alpha_j$.  \textbf{Right:} Columnar architecture in
    a convolutional network.  In this setting the $\w_\alpha$'s take linear
    combinations of the feature maps obtained by convolving the input with the
    dictionary.  We make the same abuse of notation here as in the main text---the
    vectorized filter banks must be reshaped before the convolution takes
    place.}
  \label{fig:columnar}
\end{figure}

Each $\alpha_j$ can be thought of as defining a column of representation inside
the layer.  The input to each column is shared, but the representations computed
in each column are independent.  The output of the layer is obtained by
concatenating the output of each column.  This is represented graphically in
Figure~\ref{fig:columnar}.

Introducing additional columns into the network increases the number of static
parameters but the number of dynamic parameters remains fixed.  The increase in
static parameters comes from the fact that each column has its own dictionary.
The reason that there is not a corresponding increase in the number of dynamic
parameters is that for a fixed size hidden layer the hidden units are divided
between the columns.  The number of dynamic parameters depends only on the
number of hidden units and the size of each dictionary.

In a convolutional network the interpretation is similar. In this setting we
have $g^{-1}(\h) = \v * \W^*$, where $\W^*$ is an appropriately sized filter
bank.  Using $\W$ to denote the result of vectorizing the filters of $\W^*$ (as
is done in non-convolutional models) we can again write $\W =
\U_\alpha\w_\alpha$, and using a slight abuse of notation\footnote{The
  vectorized filter bank $\W=\U_\alpha\w_\alpha$ must be reshaped before the
  convolution takes place.} we can write $g^{-1}(\h) = \v * \U_\alpha\w_\alpha$.
As above, we re-order the operations to obtain $g^{-1}(\h) =
{\v_\alpha}\w_\alpha$ resulting in a structure similar to a layer in an ordinary
MLP. This structure is illustrated in Figure~\ref{fig:columnar}.

Note that $\v$ is first convolved with $\U_\alpha$ to produce $\v_\alpha$.  That
is, preprocessing in each column comes from a convolution with a \emph{fixed}
set of filters, defined by the dictionary.  Next, we form linear combinations of
these fixed convolutions, with coefficients given by $\w_\alpha$. This
particular order of operations may result in computational improvements if the
number of hidden channels is larger than $n_\alpha$, or if the elements of
$\U_\alpha$ are separable~\cite{rigamonti2013}.

\subsection{Constructing dictionaries}
\label{ssec:construct-expanders}

We now turn our attention to selecting an appropriate dictionary for different
layers of the network.  The appropriate choice of dictionary inevitably depends
on the structure of the weight space.

When the weight space has a topological structure where we expect smoothness,
for example when the weights correspond to pixels in an image patch, we can
choose a kernel-based dictionary to enforce the type of smoothness we expect.


When there is no topological structure to exploit, we propose to use data driven
dictionaries.  An obvious choice here is to use a shallow unsupervised feature
learning, such as an autoencoder, to build a dictionary for the layer.

Another option is to construct data-driven kernels for ridge regression.  Easy
choices here are using the empirical covariance or empirical squared covariance
of the hidden units, averaged over the data.

Since the correlations in hidden activities depend on the weights in lower
layers we cannot initialize kernels in deep layers in this way without training
the previous layers.  We handle this by pre-training each layer as an
autoencoder.  We construct the kernel using the empirical covariance of the
hidden units over the data using the pre-trained weights.  Once each layer has
been pre-trained in this way we fine-tune the entire network with
backpropagation, but in this phase the kernel parameters are fixed.

We also experiment with other choices for the dictionary, such as random
projections (iid Gaussian dictionary) and random connections
(dictionary composed of random columns of the identity).


\vspace{-0.3cm}
\section{Experiments}

\vspace{-0.2cm}

\subsection{Multilayer perceptron}

\begin{figure}[t]
  \centering
  \includegraphics[width=0.46\linewidth]{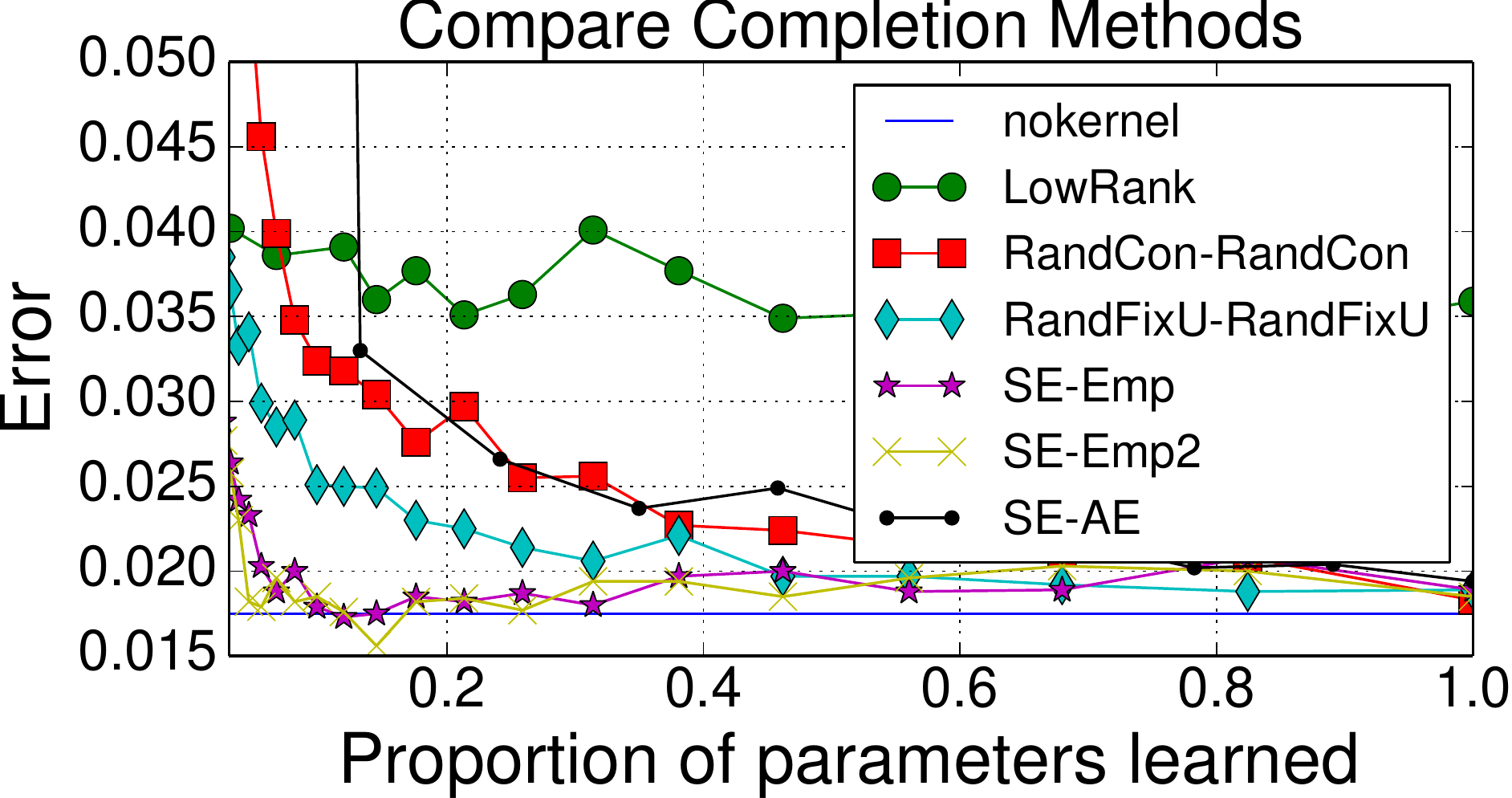}
  \qquad
  \includegraphics[width=0.46\linewidth]{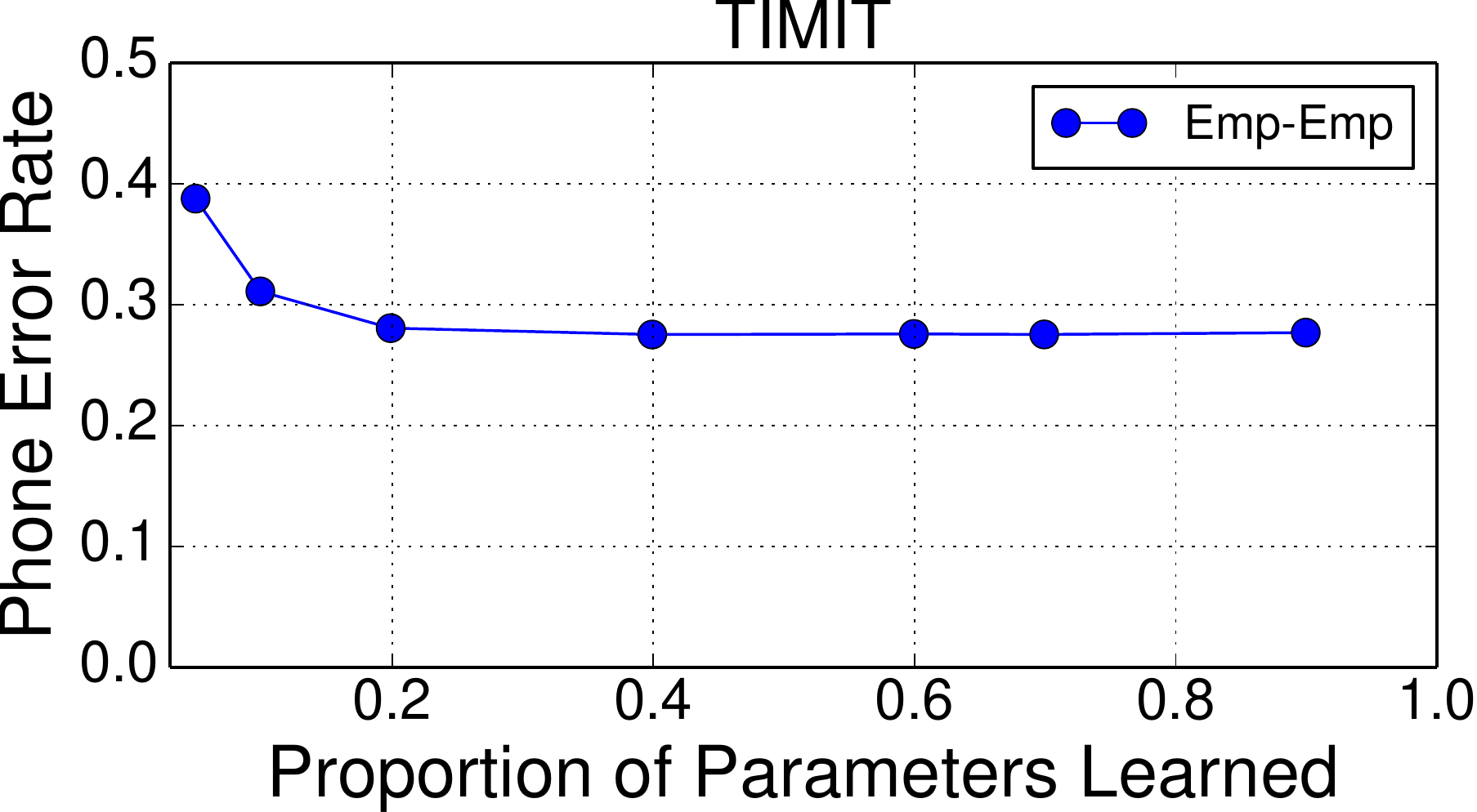}
  \caption{\textbf{Left:} Comparing the performance of different
    dictionaries when predicting the weights in the first two layers of an MLP
    network on MNIST. The legend shows the dictionary type in layer1--layer2 (see main
    text for details).  \textbf{Right:} Performance on the TIMIT core test set
    using an MLP with two hidden layers.
  }
  \label{fig:mlp}
\end{figure}

We perform some initial experiments using MLPs~\cite{rumelhart1986learning} in
order to demonstrate the effectiveness of our technique.  We train several MLP
models on MNIST using different strategies for constructing the dictionary,
different numbers of columns and different degrees of reduction in the number of
dynamic parameters used in each feature.  We chose to explore these permutations
on MNIST since it is small enough to allow us to have broad coverage.

The networks in this experiment all have two hidden layers with a
784--500--500--10 architecture and use a sigmoid activation function.  The final
layer is a softmax classifier.  In all cases we preform parameter prediction in
the first and second layers only; the final softmax layer is never predicted.
This layer contains approximately 1\% of the total network parameters, so a
substantial savings is possible even if features in this layer are not
predicted.

Figure~\ref{fig:mlp} (left) shows performance using several different strategies
for constructing the dictionary, each using 10 columns in the first and second
layers.  We divide the hidden units in each layer equally between columns (so
each column connects to 50 units in the layer above).

\begin{wrapfigure}{r}{0.35\textwidth}
  \centering
  \vspace{10pt}
  \includegraphics[width=0.96\linewidth]{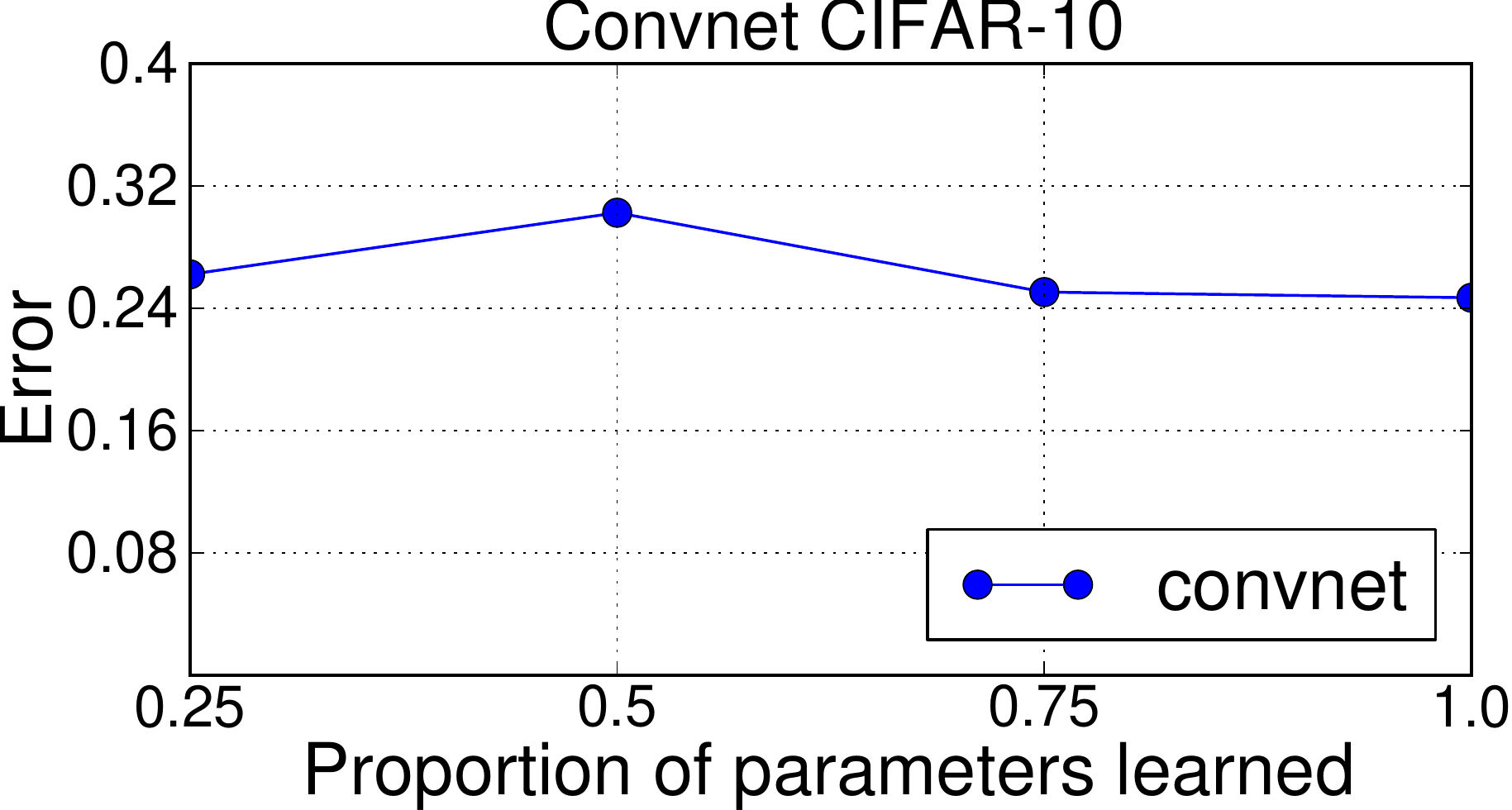}
  \caption{Performance of a convnet on CIFAR-10.  Learning only 25\% of the
    parameters has a negligible effect on predictive accuracy.}
  \label{fig:convnet}
\end{wrapfigure}

The different dictionaries are as follows: \emph{nokernel} is an ordinary model
with no feature prediction (shown as a horizontal line). \emph{LowRank} is when
both $\U$ and $\V$ are optimized. \emph{RandCon} is random connections (the
dictionary is random columns of the identity).  \emph{RandFixU} is random
projections using a matrix of iid Gaussian entries. \emph{SE} is ridge
regression with the squared exponential kernel with length scale
$1.0$. \emph{Emp} is ridge regression with the covariance kernel.  \emph{Emp2}
is ridge regression with the squared covariance kernel.  \emph{AE} is a
dictionary pre-trained as an autoencoder.  The \emph{SE--Emp} and \emph{SE-Emp2}
architectures preform substantially better than the alternatives, especially
with few dynamic parameters.

For consistency we pre-trained all of the models, except for the \emph{LowRank},
as autoencoders.  We did not pretrain the \emph{LowRank} model because we found
the autoencoder pretraining to be extremely unstable for this model.


Figure~\ref{fig:mlp} (right) shows the results of a similar experiment on TIMIT.
The raw speech data was analyzed using a 25-ms Hamming window with a 10-ms fixed
frame rate. In all the experiments, we represented the speech using 12th-order
Mel frequency cepstral coefﬁcients (MFCCs) and energy, along with their first
and second temporal derivatives.  The networks used in this experiment have two
hidden layers with 1024 units.  Phone error rate was measured by performing
Viterbi decoding the phones in each utterance using a bigram language model, and
confusions between certain sets of phones were ignored as described
in~\cite{lee1989speaker}.

\subsection{Convolutional network}


Figure~\ref{fig:convnet} shows the performance of a convnet~\cite{lecun_old} on
CIFAR-10.  The first convolutional layer filters the $32\times32\times3$ input
image using $48$ filters of size $8\times8\times3$.  The second convolutional
layer applies $64$ filters of size $8\times8\times48$ to the output of the first
layer.  The third convolutional layer further transforms the output of the
second layer by applying $64$ filters of size $5\times5\times64$.  The output of
the third layer is input to a fully connected layer with $500$ hidden units and
finally into a softmax layer with $10$ outputs.  Again we do not reduce the
parameters in the final softmax layer.  The convolutional layers each have one
column and the fully connected layer has five columns.

Convolutional layers have a natural topological structure to exploit, so we use
an dictionary constructed with the squared exponential kernel in each
convolutional layer.  The input to the fully connected layer at the top of the
network comes from a convolutional layer so we use ridge regression with the
squared exponential kernel to predict parameters in this layer as well.

\subsection{Reconstruction ICA}

Reconstruction ICA~\cite{Le-RICA} is a method for learning overcomplete ICA
models which is similar to a linear autoencoder network.  We demonstrate that we
can effectively predict parameters in RICA on both CIFAR-10 and STL-10.  In
order to use RICA as a classifier we follow the procedure of Coates et
al.~\cite{coates_analysis}.

\begin{figure}[b]
  \centering
  \includegraphics[width=0.22\linewidth]{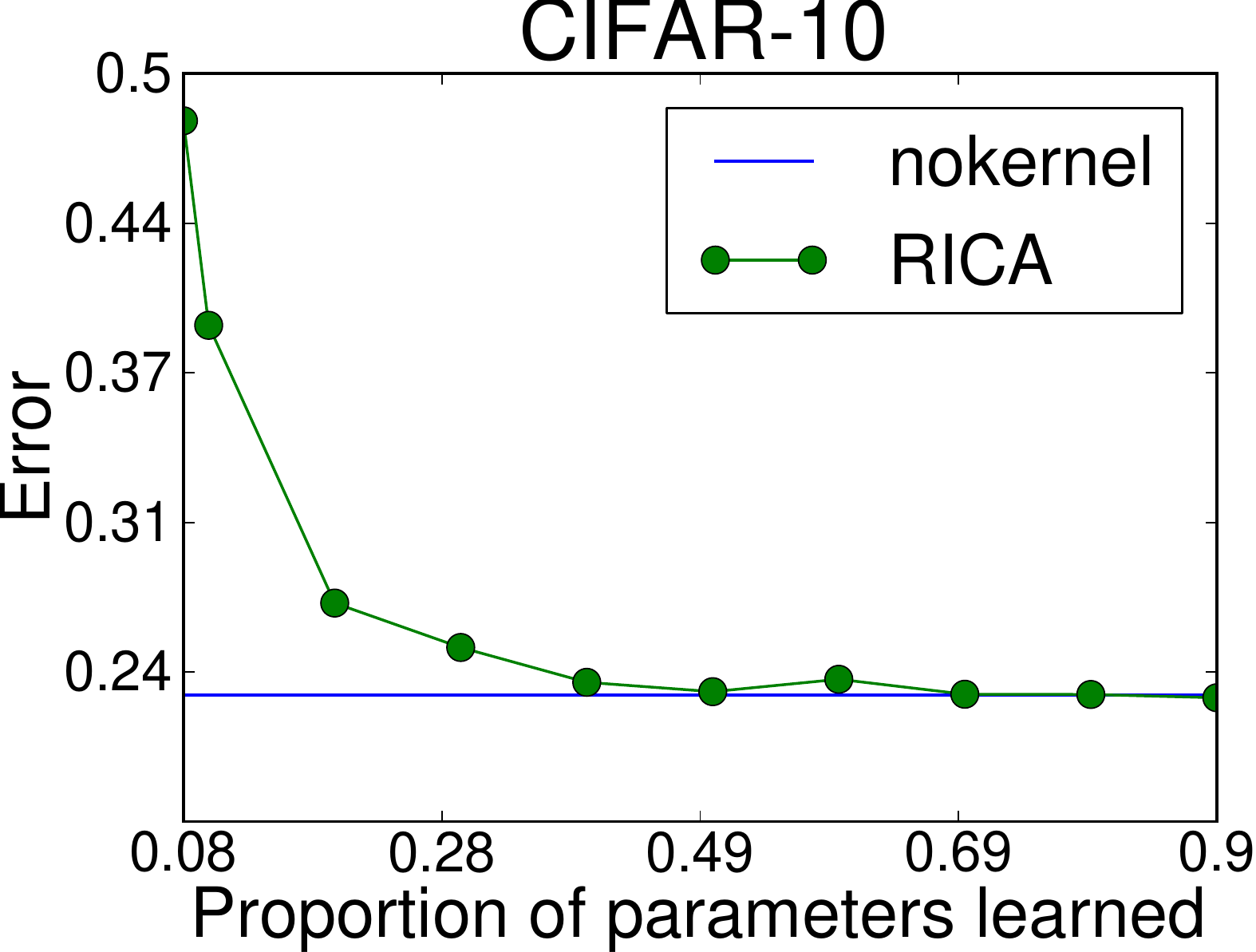}
  \quad
  \includegraphics[width=0.22\linewidth]{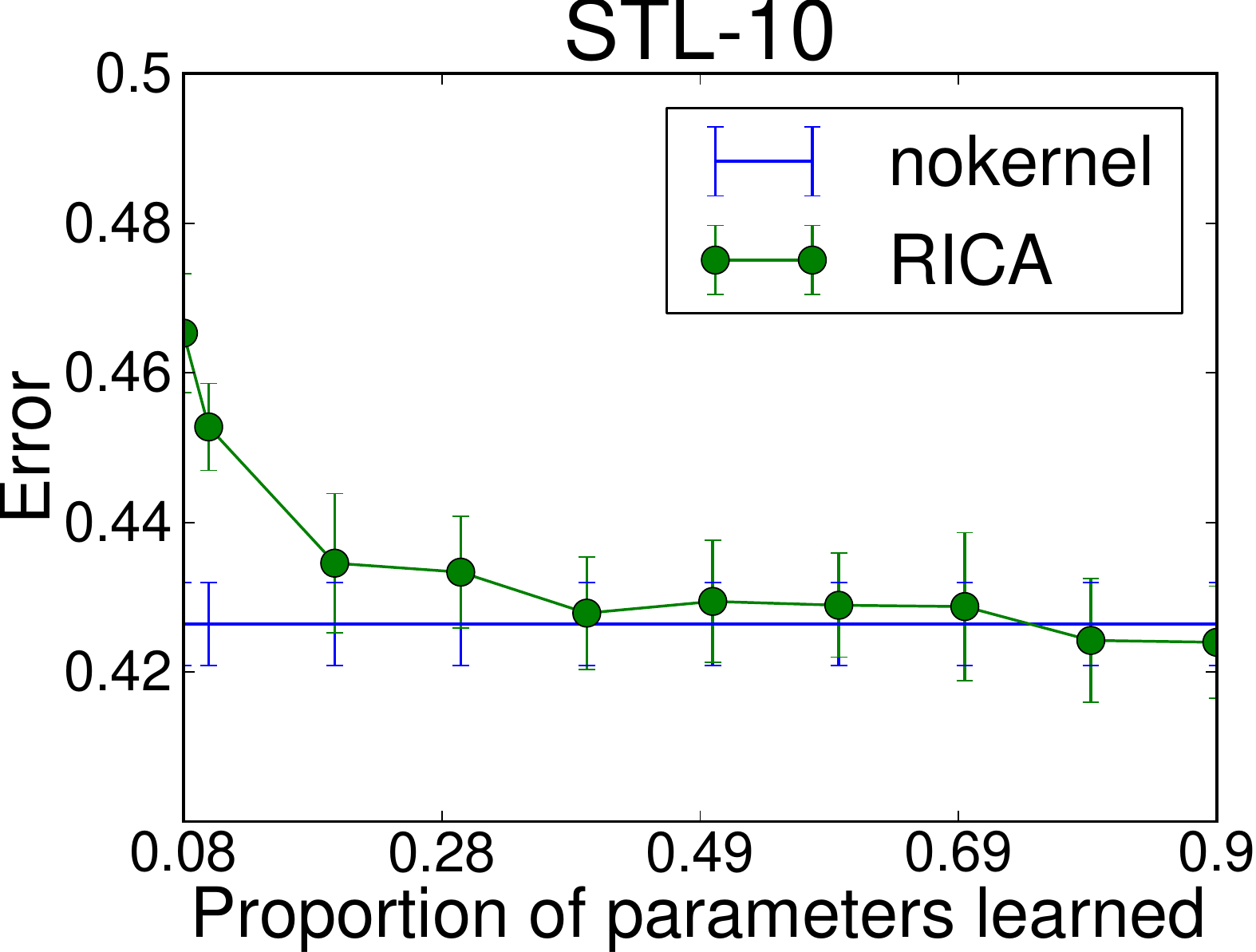}
  \qquad
  \includegraphics[width=0.22\linewidth]{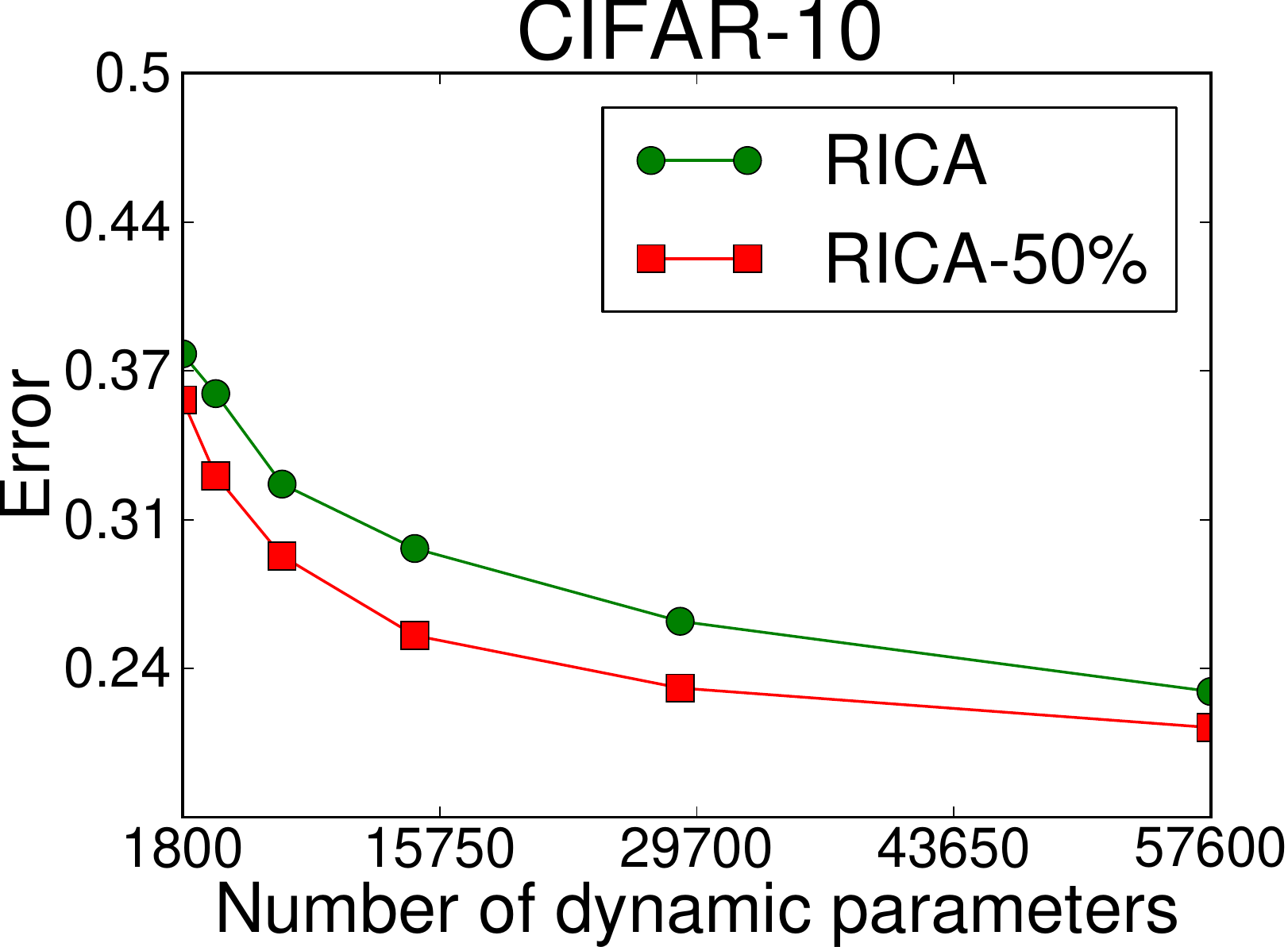}
  \quad
  \includegraphics[width=0.22\linewidth]{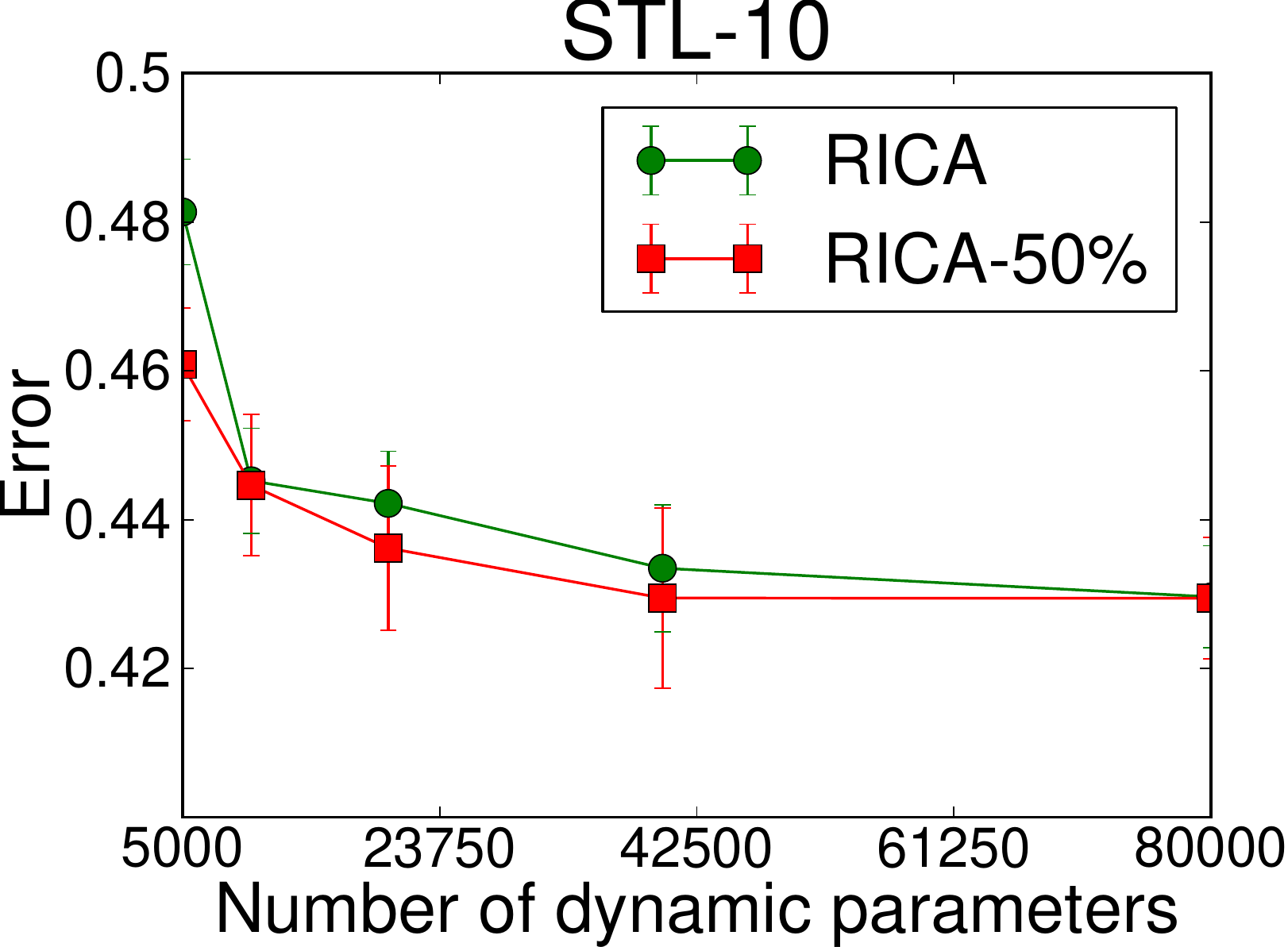}

  \caption{\textbf{Left:} Comparison of the performance of RICA with and without
    parameter prediction on CIFAR-10 and STL-10.  \textbf{Right:} Comparison of
    RICA, and RICA with 50\% parameter prediction using the same number of
    dynamic parameters (i.e.\ RICA-50\% has twice as many features).  There is a
    substantial gain in accuracy with the same number of dynamic parameters
    using our technique.  Error bars for STL-10 show 90\% confidence intervals
    from the the recommended testing protocol.  }
  \label{fig:rica}
\end{figure}

Figure~\ref{fig:rica} (left) shows the results of parameter prediction with RICA
on CIFAR-10 and STL-10.  RICA is a single layer architecture, and we predict
parameters a squared exponential kernel dictionary with a length scale of $1.0$.
The \emph{nokernel} line shows the performance of RICA with no feature
prediction on the same task.  In both cases we are able to predict more than
half of the dynamic parameters without a substantial drop in accuracy.

Figure~\ref{fig:rica} (right) compares the performance of two RICA models with
the same number of dynamic parameters.  One of the models is ordinary RICA with
no parameter prediction and the other has 50\% of the parameters in each feature
predicted using squared exponential kernel dictionary with a length scale of
$1.0$; since 50\% of the parameters in each feature are predicted, the second
model has twice as many features with the same number of dynamic parameters.


\section{Related work and future directions}

Several other methods for limiting the number of parameters in a neural network
have been explored in the literature.  An early approach is the technique of
``Optimal Brain Damage''~\cite{lecun90} which uses approximate second derivative
information to remove parameters from an already trained network.  This
technique does not apply in our setting, since we aim to limit the number of
parameters before training, rather than after.

The most common approach to limiting the number of parameters is to use locally
connected features~\cite{coates_analysis}.  The size of the parameterization of
locally connected networks can be further reduced by using tiled convolutional
networks~\cite{gregor2010emergence} in which groups of feature weights which
tile the input space are tied.  Convolutional neural
networks~\cite{alex_imagenet} are even more restrictive and force a feature to
have tied weights for all receptive fields.

Techniques similar to the one in this paper have appeared for shallow models in
the computer vision literature.  The double sparsity method of Rubinstein et
al.~\cite{rube2010} involves approximating linear dictionaries with other
dictionaries in a similar manner to how we approximate network features.
Rigamonti et al.~\cite{rigamonti2013} study approximating convolutional filter
banks with linear combinations of separable filters.  Both of these works focus
on shallow single layer models, in contrast to our focus on deep networks.

The techniques described in this paper are orthogonal to the parameter reduction
achieved by tying weights in a tiled or convolutional pattern.  Tying weights
effectively reduces the number of feature maps by constraining features at
different locations to share parameters.  Our approach reduces the number of
parameters required to represent each feature and it is straightforward to
incorporate into a tiled or convolutional network.

Cire\c{s}an \emph{et al.}~\cite{ciresan2011} control the number of parameters by
removing connections between layers in a convolutional network at random. They
achieve state-of-the-art results using these randomly connected layers as part
of their network.  Our technique subsumes the idea of random connections, as
described in Section~\ref{ssec:construct-expanders}.

The idea of regularizing networks through prior knowledge of smoothness is not
new, but it is a delicate process.  Lang and Hinton~\cite{hinton1990} tried
imposing explicit smoothness constraints through regularization but found it to
universally reduce performance.  Na\"ively factoring the weight matrix and
learning both factors tends to reduce performance as well.  Although the idea is
simple conceptually, execution is difficult.  G\"ul\c{c}ehre et
al.~\cite{gulcehre2013} have demonstrated that prior knowledge is extremely
important during learning, which highlights the importance of introducing it
effectively.

Recent work has shown that state of the art results on several benchmark tasks
in computer vision can be achieved by training neural networks with several
columns of representation~\cite{ciresan2012multi,alex_imagenet}.  The use of
different preprocessing for different columns of representation is of particular
relevance~\cite{ciresan2012multi}.  Our approach has an interpretation similar
to this as described in Section~\ref{ssec:columns}.  Unlike the work
of~\cite{ciresan2012multi}, we do not consider deep columns in this paper;
however, collimation is an attractive way for increasing parallelism within a
network, as the columns operate completely independently.  There is no reason we
could not incorporate deeper columns into our networks, and this would make for
a potentially interesting avenue of future work.

Our approach is superficially similar to the factored
RBM~\cite{marc2010factored,ICML2011Swersky}, whose parameters form a 3-tensor.
Since the total number of parameters in this model is prohibitively large, the
tensor is represented as an outer product of three matrices.  Major differences
between our technique and the factored RBM include the fact that the factored
RBM is a specific model, whereas our technique can be applied more
broadly---even to factored RBMs.  In addition, in a factored RBM all factors are
learned, whereas in our approach the dictionary is fixed judiciously.


In this paper we always choose the set $\alpha$ of indices uniformly at random.
There are a wide variety of other options which could be considered here.  Other
works have focused on learning receptive fields directly~\cite{coates11}, and
would be interesting to incorporate with our technique.

In a similar vein, more careful attention to the selection of kernel functions
is appropriate.  We have considered some simple examples and shown that they
preform well, but our study is hardly exhaustive.  Using different types of
kernels to encode different types of prior knowledge on the weight space, or
even learning the kernel functions directly as part of the optimization
procedure as in~\cite{vincent2000neural} are possibilities that deserve
exploration.

When no natural topology on the weight space is available we infer a topology
for the dictionary from empirical statistics; however, it may be possible to
instead construct the dictionary to induce a desired topology on the weight
space directly.  This has parallels to other work on inducing topology in
representations~\cite{gregor2010emergence} as well as work on learning pooling
structures in deep networks~\cite{coates2012emergence}.


\vspace{-5pt}
\section{Conclusion}
\vspace{-3pt}

We have shown how to achieve significant reductions in the number of dynamic
parameters in deep models. The idea is orthogonal but complementary to recent
advances in deep learning, such as dropout, rectified units and maxout. It
creates many avenues for future work, such as improving large scale industrial
implementations of deep networks, but also brings into question whether we have
the right parameterizations in deep learning.


\clearpage
\small{
\subsubsection*{References}
\let\u\uaccent
\bibliographystyle{plainabbrv}
\bibliography{refs}
}

\end{document}